\documentclass[a4paper, 10pt, journal]{IEEEtran}      
\IEEEoverridecommandlockouts                              
\usepackage{adjustbox}
\usepackage{graphicx} 
\usepackage{epsfig} 
\usepackage{amsmath, amssymb} 
\usepackage{booktabs}
\usepackage{siunitx}
\usepackage{soul,framed}
\usepackage{color}
\usepackage[table]{xcolor}
\usepackage{arydshln}
\usepackage[flushleft]{threeparttable}

\definecolor{darkgreen}{rgb}{0,0.6,0.2}

\begin{document}

\title{\LARGE \bf
ROAD MAPPING IN LIDAR IMAGES USING A JOINT-TASK DENSE DILATED CONVOLUTIONS MERGING NETWORK
}


\author{Qinghui Liu$^{1,2}$,~\IEEEmembership{Student Member,~IEEE,} Michael Kampffmeyer$^{2}$,~\IEEEmembership{Student Member,~IEEE,} Robert Jenssen$^{2,1}$,~\IEEEmembership{Member,~IEEE,} and Arnt-B{\o}rre Salberg$^{1}$,~\IEEEmembership{Member,~IEEE}\\  
\thanks{$^{1}$Norwegian Computing Center, Dept. SAMBA, P.O. Box 114 Blindern, NO-0314 OSLO, Norway}%
\thanks{$^{2}$UiT Machine Learning Group, Department of Physics and Technology, UiT the Arctic University of Norway, Troms{\o}, Norway
        }%
}

\maketitle

\thispagestyle{empty}
\pagestyle{empty}

\setlength{\textfloatsep}{8pt plus 0pt minus 5pt}

\begin{abstract}
It is important, but challenging, for the forest industry to accurately map roads which are used for timber transport by trucks. In this work, we propose a Dense Dilated Convolutions Merging Network (DDCM-Net) to detect these roads in lidar images. The DDCM-Net can effectively recognize multi-scale and complex shaped roads with similar texture and colors, and also is shown to have superior performance over existing methods. To further improve its ability to accurately infer categories of roads, we propose the use of a joint-task learning strategy that utilizes two auxiliary output branches, i.e, multi-class classification and binary segmentation, joined with the main output of full-class segmentation. This pushes the network towards learning more robust representations that are expected to boost the ultimate performance of the main task. 
In addition, we introduce an iterative-random-weighting method to automatically weigh the joint losses for auxiliary tasks. This can avoid the difficult and expensive process of tuning the weights of each task's loss by hand. The experiments demonstrate that our proposed joint-task DDCM-Net can achieve better performance with fewer parameters and higher computational efficiency than previous state-of-the-art approaches. 
\end{abstract}



\section{Introduction}
Automatic detection and mapping of road networks from remote sensing data has been previously studied extensively, however, most of the works focus on optical data and many algorithms fail to extract roads well in optical images for cases where surrounding objects like water, trees, grass, and buildings occlude the road \cite{wang2016review}.
In recent years, a large variety of modern approaches to pixel classification and segmentation are based on deep convolutional neural networks (CNN) \cite{kampffmeyer2016semantic, paisitkriangkrai2015effective, Sherrah16, audebert2016semantic, MarmanisSWGDS16, wang2017gated}, in particular end-to-end learning with fully convolutional neural networks (FCN) \cite{long2015fully}. However, to achieve higher performance, FCN-based methods normally rely on deep multi-scale architectures which typically require a large number of trainable parameters and computation resources. 

 \begin{figure}[tbp]
 \centering
  \includegraphics[width=0.47\textwidth]{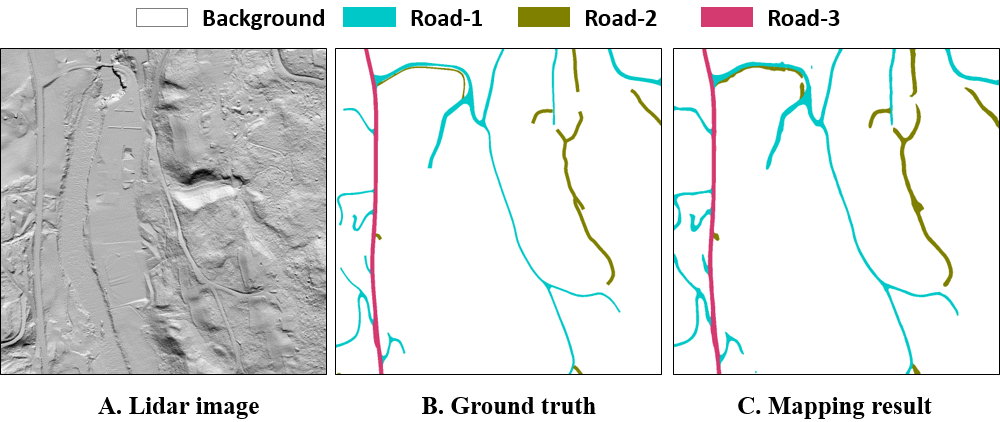} 
  \caption{Forest roads mapping in a lidar image with our Joint-task DDCM network.}
  \label{fig:semapping}
\end{figure}


In this work, we propose a joint-task learning method with a novel network architecture, called the dense dilated convolutions merging network (DDCM-Net), which utilizes multiple dilated convolutions merged with various dilation rates. The proposed network learns with densely linked dilated convolutions and outputs a fusion of all intermediate features during the extraction of multi-scale features. Our experiments demonstrate that the network combined with our joint-task learning strategy achieves robust and accurate results with relatively few parameters and layers. Fig. \ref{fig:semapping} illustrates an example of roads mapping results on lidar data with the joint-task DDCM-Net. These results will be further discussed in Section \ref{exp}.

\section{Methods}
We first briefly revisit dilated convolutions, which are used in DDCM networks. We then present our DDCM architecture with the joint-task learning method and further provide training details.

\subsection{Dilated Convolutions}
Dilated convolutions~\cite{yu2015multi} allow us to flexibly adjust the filter's receptive field to capture multi-scale information without increasing the number of parameters. A 2-D dilated convolution operator can be defined as

\begin{equation}\label{dconv} 
   g_{i, j}(F_{i-1}) = \sum_{k=0}^{c_{i-1}}  W_{h_{ijk}, r} \ast F_{i-1}^{k}
\end{equation}
where, $\ast$ denotes a convolution operator, $g_{i,j}: \mathbb{R}^{m_{i-1} \times n_{i-1} \times  c_{i-1}} \rightarrow \mathbb{R}^{m_{i} \times n_{i}}$ convolves each channel of the input feature map  $F_{i} \in \mathbb{R}^{m_{i} \times n_{i} \times  c_{i}}$, $m$ and $n$ denote the spatial dimensions and $c$ the number of channels. A 2-D dilated convolution $W_{h, r}$ with a filter $h$ and dilation $r \in \mathbb{Z}^{+}$ is only nonzero for a multiple of $r$ pixels from the center. In dilated convolution, a kernel $k\times k$ is enlarged to $k+(k-1)(r-1)$ with the dilation factor $r$. As a special case, a dilated convolution with dilation rate $r=1$ corresponds to a standard convolution. 
 \begin{figure}[htbp]
 \centering
  \includegraphics[width=0.48\textwidth]{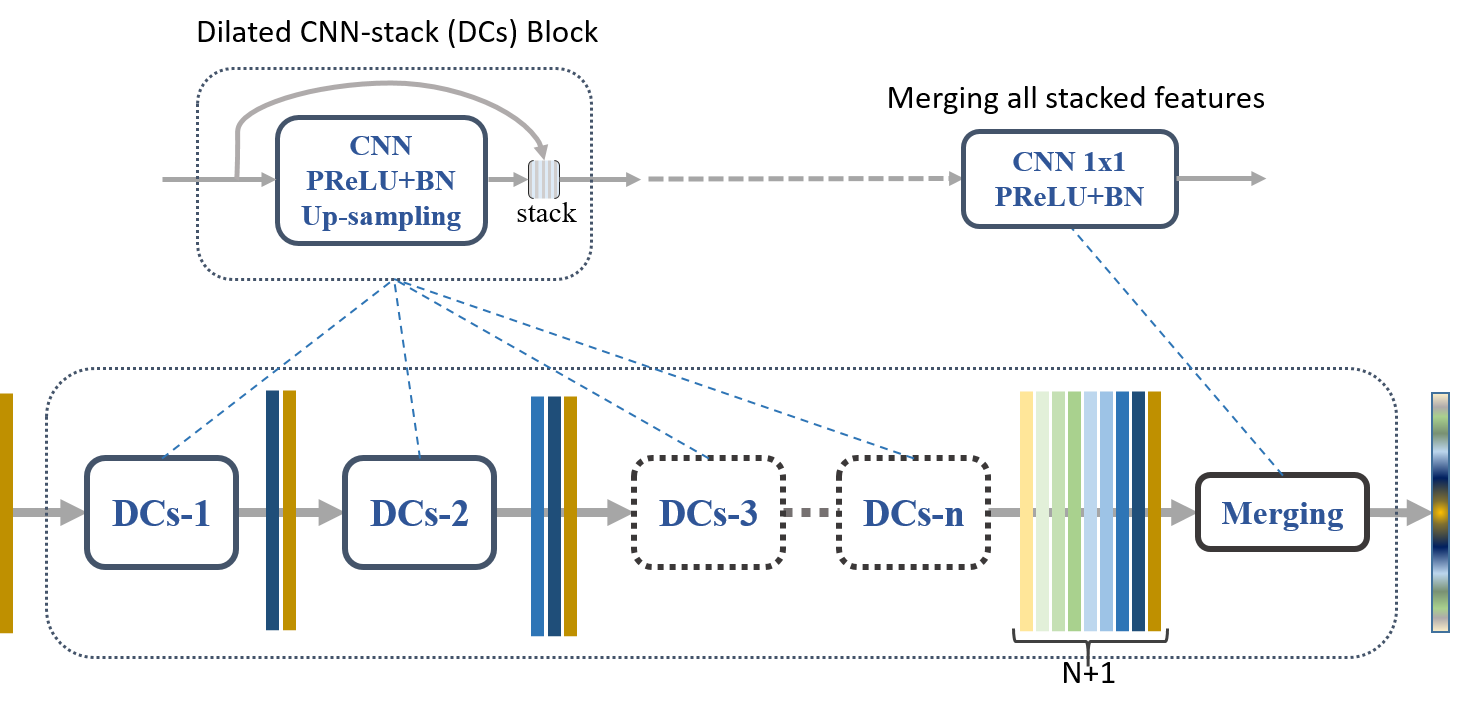} 
  \caption{Example of the DDCM-Net architecture composed of $N$ DC modules with various dilation rates $\{1, 2, 3, ... , N\}$. A basic DCs block is composed of a dilated convolution with a stride of 2, followed by PReLU \cite{he2015delving}, batch normalization and up-sampling. It then stacks the output with its input together to feed the next layer.}
  \label{fig:ddcm}
\end{figure}

 \begin{figure*}[htbp]
 \centering
  \includegraphics[width=0.78\textwidth]{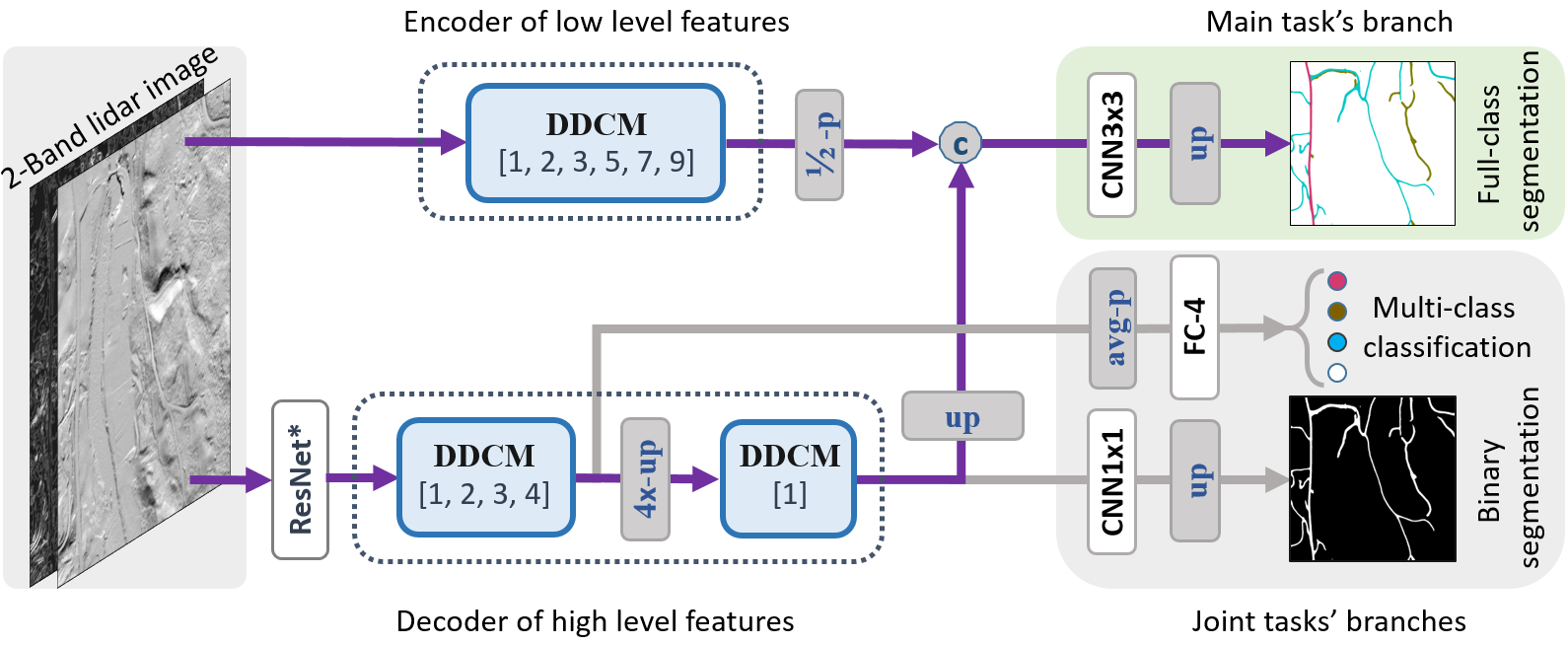} 
  \caption{End-to-end pipeline of the Joint-Task DDCM-Net (JT-DDCM-R50) for roads mapping in lidar images. The encoder of low level features encodes multi-scale contextual information from the initial 2-band lidar images by a DDCM module with 6 different dilation rates $[1, 2, 3, 5, 7, 9]$. The decoder of high level features decodes highly abstract representations learned from ResNet by 2 DDCM modules with rates $[1, 2, 3, 4]$ and $[1]$ separately. The transformed low-level and high-level feature maps by DDCMs are then fused together to infer pixel-wise full-class probabilities. There are also a multi-class classification output which predicts what types of roads are in the input, and a binary segmentation output which locates all roads. Whereby, we call this design a joint-task learning model. Here, 'p', 'up', 'c', 'avg-p' and 'FC' denote max-pooling, up-sampling, concatenation, adaptive average pooling and fully connected output respectively.}
  \label{fig:ddcm_resnet}
\end{figure*}

\subsection{DDCM-Net}

Dense Dilated Convolutions Merging Network (DDCM-Net) consists of a number of Dilated CNN-stack (DCs) blocks with a merging module as output. A basic DCs block is composed of a dilated convolution using a stride of 2 to reduce computational cost, followed by PReLU \cite{he2015delving} non-linear activation, batch normalization (BN) \cite{ioffe2015batch} and bilinear up-sampling to scale the output to be the same size as the input. It then stacks the output with its input together to feed to the next layer, which can alleviate loss of context information and problems with vanishing gradients when adding more layers. The final network output is computed by a merging module composed of $1\times1$ filters with BN and PReLU in order to efficiently combine all stacked features generated by intermediate DCs blocks. In practice, densely connected DCs blocks, typically configured with progressively or exponentially increasing dilation factors, enable DDCM networks to have very large receptive fields with just a few layers as well as to capture scale-invariant features by merging multi-scale features properly. Fig. \ref{fig:ddcm} illustrates the basic structure of the DDCM network. 
\subsection{Joint-Task with Iterative-Random-Weighting Losses}
Fig. \ref{fig:ddcm_resnet} shows the end-to-end pipeline of a joint-task DDCM-Net combined with a tailored ResNet pre-trained on ImageNet \cite{ILSVRC15} for road mapping tasks. We only utilize the first 3 bottleneck layers of ResNet50 and remove the last bottleneck layer and fully connected layers to reduce the number of trainable parameters. 
For cases when the input data consists of less than three bands, we add a standard convolution with $3 \times 3$ kernels followed by BN and PReLU on top of the ResNet in order to compute the remaining bands. We then stack the generated bands with the input image, forming a 3-band input image. 

We apply a 2D cross-entropy loss function with median frequency balancing as defined in \cite{kampffmeyer2016semantic} as the main task loss ($L_{mce}$), a binary cross entropy (BCE) loss ($L_{bce}$) and a Lov{\'a}sz-softmax loss \cite{lovasz} ($L_{lovasz}$) for the two joint tasks of multi-class classification and binary segmentation separately. In addition, to avoid the difficult and expensive process of tuning weights of the joint-task's losses by hand, we introduce an iterative-random-weighting method to randomly weight the joint-loss for each iteration during a training epoch. The total loss ($L_{total}$) will sum the main loss with automatically weighted joint losses. We sample the weights $w_{r1, i}$ and $w_{r2, i}$ uniformly between 0 and 1. 
The iterative-random-weighting loss can therefore be defined as
\begin{equation}\label{loss} 
   L_{total}^i = L_{mce}^i + w_{r1, i} \ast L_{bce}^i + w_{r2, i} \ast L_{lovasz}^i
\end{equation}
where, $i \in \{1, 2, 3, ...\}$ denotes the current training iteration. 

\subsection{Optimizer and Multi-Step LR policy}
In our work, we choose Adam \cite{KingmaB14adam} with AMSGrad \cite{amsgrad2018} as the optimizer for the model. Guided by our empirical results, we utilized multi-step learning rate (LR) schedule method. The multi-Step LR policy drops the learning rate by 0.5 at epochs $[5, 15, 25, 65, 100]$ with initial LR $0.00012$ and iterative weight decay $0.00005$. We also set $2 \times LR$ to all bias parameters in contrast to weights parameters.

\section{Experiments}
\label{exp}
We first investigate the proposed DDCM-Net (DDCM-R50 model) on the publicly available ISPRS 2D semantic labeling contest datasets \cite{ISPRS2018}, which contains two state-of-the-art airborne image datasets (Potsdam and Vaihingen). We compare with related published or re-implemented state-of-the-art methods \cite{liuqh2018}. We further evaluate the DDCM-R50 model on our own lidar dataset with and without joint-task learning separately, and discuss the results and compare with our previous work \cite{salberg2017large} as well. Please note that the structure of DDCM-R50 model for ISPRS datasets is just slightly different with the one for lidar dataset which has an extra input layer on top of ResNet50. The extra input layer uses one standard convolution with $3 \times 3$ kernels to compute the third band which is then stacked with the original 2-band input forming a 3-band input for the pre-trained ResNet.

\subsection{Lidar Dataset and Metrics}
 The lidar dataset is composed of 2 very high resolution ($19200 \times 12800$) images that contain 2 bands: an elevation gradient band and a hillshade band \cite{salberg2017large}. The data is annotated with 4 classes including background, Road1, Road2, and Road3. To evaluate our models, one lidar image was divided into a training and validation dataset and one was used for testing. The performance is measured on the test set by both the F1-score \cite{kampffmeyer2016semantic}, and the mean Intersection over Union (IoU) \cite{liuqh2018}. 

\subsection{Train and Test Time Augmentation}
We randomly sample 5000 image patches ($256 \times 256$) for the ISPRS dataset and 1000 patches ($1024 \times 1024$) for the lidar data respectively in run time from the training images for each training epoch and flip or mirror images for data augmentation. These patches are normalized to [0.0, 1.0]. No mean and standard deviation normalization is used. 

We also apply test time augmentation (TTA) with flipping and mirroring during testing period. We use sliding windows (with approximately 60\% overlap) on a test image and stitched the results together by averaging the predictions of the over-lapping TTA regions to form the whole mapping output. 
Please note that we use different windows' sizes, ($448 \times 448$) for ISPRS and ($1024 \times 1024$) for Lidar data respectively. 

\subsection{Results and Discussions}

\begin{table}[hptb!]
\centering 
  \caption{Quantitative Comparison of parameters size, FLOPs (measured on input image size of $3 \times 256 \times 256$), and mIoU  on ISPRS Potsdam RGB dataset.}
\resizebox{0.95\columnwidth}{!}{
\begin{threeparttable}
\begin{tabular}{c||p{18mm}p{18mm}p{18mm}} \hline 
\textbf{Models} \cite{liuqh2018} & \textbf{Parameters} \newline (Million)& \textbf{FLOPs} \newline (Giga) & \textbf{mIoU} \cite{liuqh2018} \\  \hline
UNet-VGG16 & 31.04 & 15.25 & 0.715 \\
FCN8s-VGG16 & 134.30 & 73.46 & 0.728 \\
SegNet-VGG19 & 39.79 & 60.88 & 0.781 \\
GCN-ResNet50 & 23.84 & 5.61  & 0.774 \\
PSP-ResNet50 & 46.59 & 44.40 & 0.789 \\
DUC-ResNet50 & 30.59 & 32.26 & 0.793 \\ \hdashline
DDCM-R50 & \cellcolor{gray!25}\textbf{9.99}  & \cellcolor{gray!25}\textbf{4.86} &\cellcolor{gray!25} \textbf{0.808} \\ \hdashline
JT-DDCM-R50 & 9.30$^{*}$  & 4.44$^{*}$ & - \\ \hline
\end{tabular}
    \begin{tablenotes}
            \item[*] It is measured on 2-band ($2 \times 256 \times 256$) input data.
    \end{tablenotes}
  \end{threeparttable}
}
\label{tab:parameters}%
\end{table}

\begin{table}[hptb!]
\centering 
  \caption{Comparisons between DDCM-R50 with other published methods on ISPRS Vaihingen IRRG dataset. }
\resizebox{\columnwidth}{!}{
\begin{threeparttable}
\begin{tabular}{c|p{8mm}|p{12mm}p{9mm}p{11mm}p{9mm}p{8mm}|p{8mm}} \hline
    \textbf{Models} & $\textbf{OA}$ & \textbf{Buildings} & \textbf{Trees} & \textbf{Low-veg} & \textbf{Surfaces} & \textbf{Cars}  & \textbf{mF1} \\  \hline \hline
    ADL\_3 \cite{paisitkriangkrai2015effective} & 0.880  & 0.932  & 0.882  & 0.823 & 0.895  & 0.633  & 0.833 \\ 
    DST\_2 \cite{Sherrah16} & 0.891  & 0.937  & 0.892  & 0.834 & 0.905  & 0.726 & 0.859 \\ 
    ONE\_7 \cite{audebert2016semantic} & 0.898  & 0.945  & 0.899  & \cellcolor{gray!25}\textbf{0.844} & 0.910  & 0.778 & 0.875\\ 
    DLR\_9 \cite{MarmanisSWGDS16} & 0.903  & 0.952  & 0.899  & 0.839 & 0.924  & 0.812  & 0.885 \\  
    GSN \cite{wang2017gated} & 0.903  & 0.951  & \cellcolor{gray!25}\textbf{0.899}  & 0.837 & 0.922  & 0.824  & 0.887 \\\hdashline
    DDCM-R50$^{*}$ (ours) & \cellcolor{gray!25}\textbf{0.904} \newline (+0.1\%) & \cellcolor{gray!25}\textbf{0.953} \newline (+0.1\%)   &0.894 \newline (-0.5\%)  & 0.833 \newline (-1.1\%)  & \cellcolor{gray!25}\textbf{0.927}\newline (+0.3\%) & \cellcolor{gray!25}\textbf{0.883}\newline (+5.9\%) & \cellcolor{gray!25}\textbf{0.898} \newline (+1.1\%)\\  \hline
\end{tabular}
\begin{tablenotes}
            \item[*] Evaluated on the hold-out test set with only IRRG bands, which contains 17 IRRG images of tiles ID 2, 4, 6, 8, 10, 12, 14, 16, 20, 22, 24, 27, 29, 31, 33, 35 and 38. 
        \end{tablenotes}
  \end{threeparttable}
} 
\label{tab:vaihingen}%
\end{table}

\begin{figure}[htpb!]
 \centering
  \includegraphics[width=0.48\textwidth]{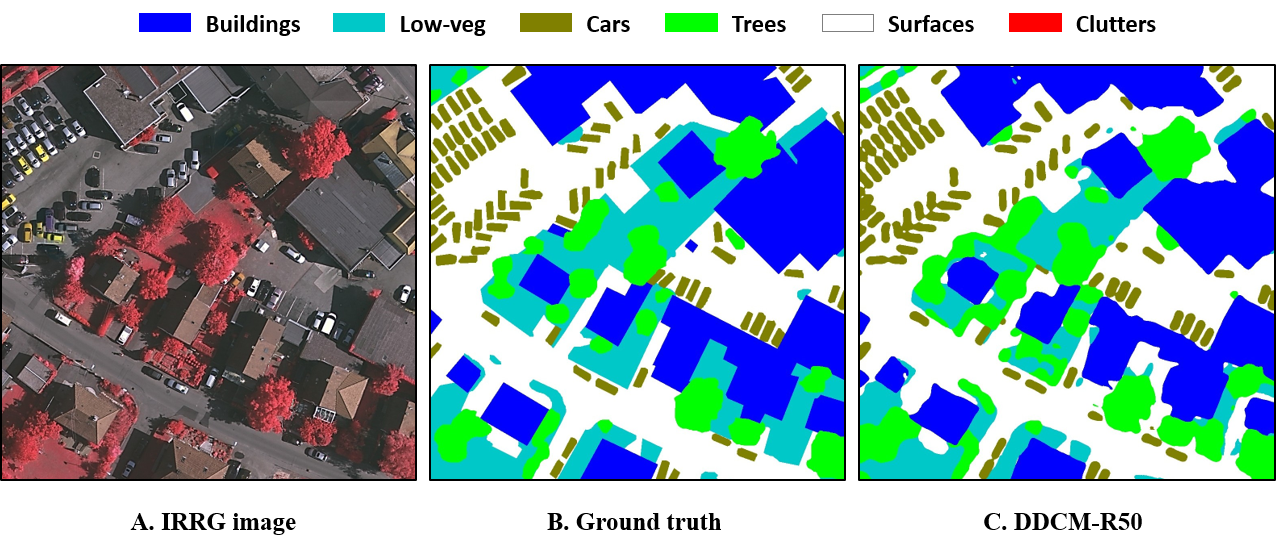} 
  \caption{Mapping results for an image patch ($1024 \times 1024$) of tile-27 IRRG test image with single DDCM-R50 model.}
  \label{fig:irrg_test}
\end{figure}

Table \ref{tab:parameters} and \ref{tab:vaihingen} show our test results of DDCM-R50 on the ISPRS Potsdam RGB images and Vaihingen IRRG images separately. Please note that the mode was evaluated with full-reference boundary ground truths on Potsdam RGB data, and compared to our re-implementation of other popular methods \cite{liuqh2018}, while for the Vaihingen IRRG data, it was evaluated with eroded boundary ground truths in order to fairly compare with other published methods. 

Our single DDCM-R50 model achieved the highest average IoU score (80.8\%) compared to other popular architectures \cite{liuqh2018} on Potsdam RGB dataset, while using more than 4 times fewer parameters than PSP-ResNet50. Please note that our IoU was eveluated on full reference ground truths. And more, our model only required 4.86 GFLOPs (measured on input size of $3\times256\times256$), which also outperformed the state of the art models in terms of computational efficiency. On Vaihingen IRRG dataset, our model also obtained the best overall accuracy and F1-score, which is +1.1\% higher mF1 than the second best model. Fig. \ref{fig:irrg_test} shows a qualitative mapping results on a Vaihingen IRRG test image with our model. 


\begin{table}[hptb!]
\centering 
  \caption{Full-class Mapping Results on the test lidar images with the joint-task DDCM (JT-DDCM-R50) model, here BG denotes background.}
\resizebox{0.75\columnwidth}{!}{
\begin{tabular}{c|c|cccc} 
    \textbf{Models} & \textbf{mF1} & \textbf{BG} & \textbf{Road1} & \textbf{Road2} & \textbf{Road3}  \\ \hline 
    DDCM-R50 & 0.650  & 0.995  & 0.619 & \cellcolor{gray!25}\textbf{0.284}  & 0.702 \\ 
    \hdashline
    \cellcolor{gray!25}JT-DDCM-R50 & \cellcolor{gray!25}\textbf{0.696}  & \cellcolor{gray!25}\textbf{0.997}  & \cellcolor{gray!25}\textbf{0.715} & 0.282  & \cellcolor{gray!25}\textbf{0.789}  \\  \bottomrule 
    \textbf{} & \textbf{mIoU} & \textbf{} & \textbf{} & \textbf{} & \textbf{}  \\  \hline
    DDCM-R50 & 0.536 & 0.991 & 0.448 & \cellcolor{gray!25}\textbf{0.165} & 0.541  \\  \hdashline
    \cellcolor{gray!25}JT-DDCM-R50 & \cellcolor{gray!25}\textbf{0.592}  & \cellcolor{gray!25}\textbf{0.994}  & \cellcolor{gray!25}\textbf{0.556} & 0.164  & \cellcolor{gray!25}\textbf{0.652}  \\ \hline 
\end{tabular}
}
\label{tab:roadresults}%
\end{table}
 
\begin{figure}[htpb!]
 \centering
  \includegraphics[width=0.48\textwidth]{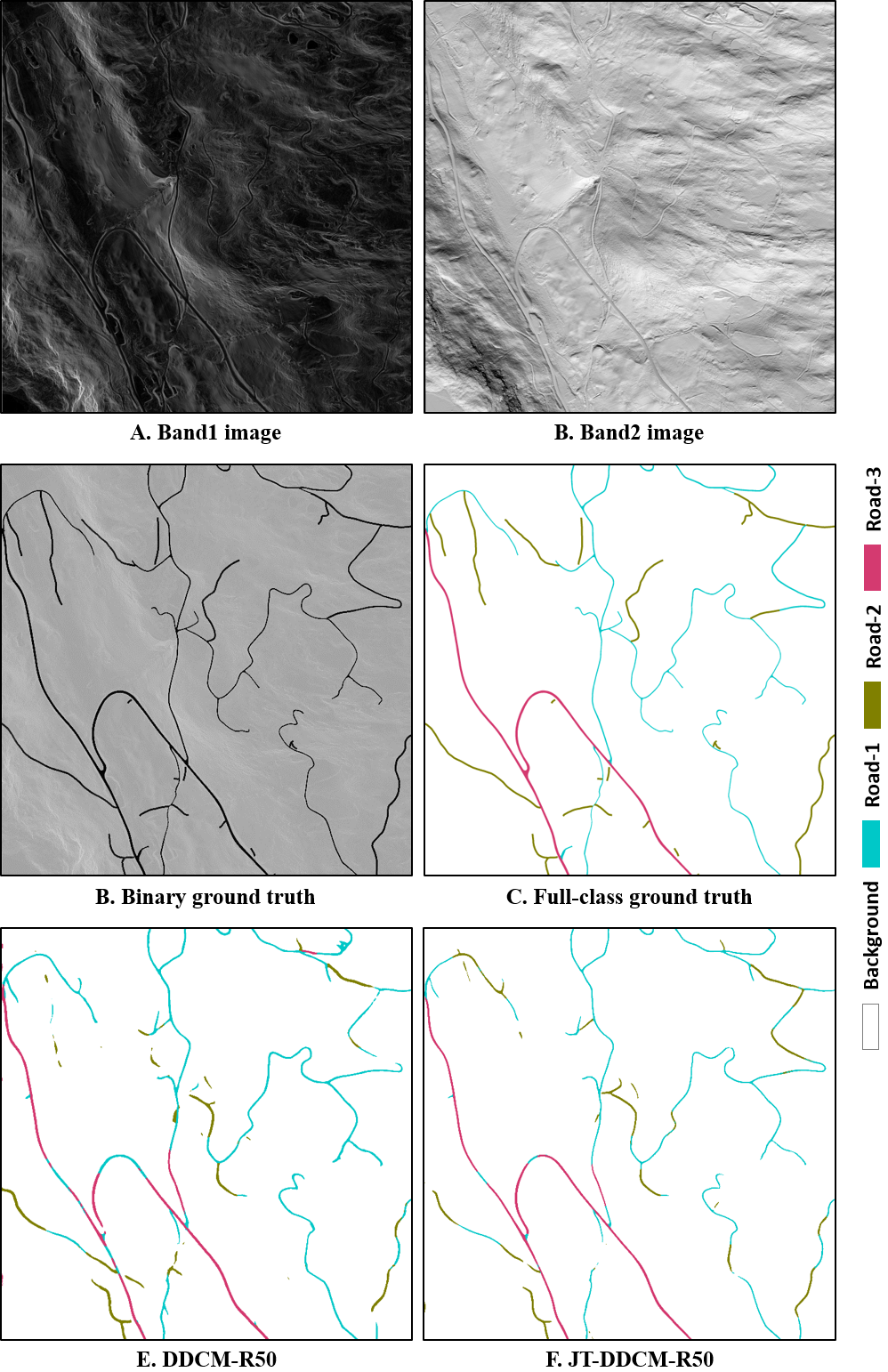} 
  \caption{Mapping results of 2-band lidar data patch ($2500 \times 2500$) with DDCM-R50 and JT-DDCM-R50 respectively.}
  \label{fig:roadtest}
\end{figure}

Table \ref{tab:roadresults} presents the road mapping results of the DDCM-R50 and JT-DDCM-R50 models on the lidar test set, respectively. The joint-task model JT-DDCM-R50 obtained average IoU ($0.592$) and F1-score ($0.696$), which are $5.6\%$ and $4.6\%$ higher than DDCM-R50 model. Both models achieved an overall accuracy above $99.3\%$, which is about $2.1\%$ higher than the previous work \cite{salberg2017large}. Fig. \ref{fig:roadtest} shows a qualitative comparison of the road mapping results. In general, the JT-DDCM-R50, which utilized joint-task strategies, obtained less fragmented mapping than the original DDCM-R50 model without joint-task learning. However, both models tend to easily mis-classify  Road2 as Road1. From the ground truth data, we observe that the two types of roads, Road1 and Road2, have highly similar characteristics in both the gradient and hillshade bands. We think that adding more training data of Road2 samples could improve the performance.

\section{Conclusions}
In this paper, we presented a dense dilated convolutions merging (DDCM) network architecture and a joint-task learning structure with a iterative-random-weighting strategy for the joint-loss. By applying dilated convolutions to learn features at varying dilation rates and merging the feature map of each layer with the feature maps from all previous layers, the DDCM-Net architecture can achieve competitive results with much fewer parameters and more computational efficiency than existing architectures. DDCM-Net is easy to implement, train and combine with existing architectures to address a wide range of different problems, and the proposed joint-task learning framework further boosts performance.


\section*{ACKNOWLEDGMENT}
This work is supported by the foundation of the Research Council of Norway under Grant 220832 and Grant 239844. Hamar regional office also provided vector data. Airborne laser scanning data was provided by Oppland Country Administration.



\begin{thebibliography}{10}

\bibitem{wang2016review}
W.~Wang, N.~Yang, Y.~Zhang, F.~Wang, T.~Cao, and P.~Eklund, ``A review of road
  extraction from remote sensing images,'' {\em Journal of traffic and
  transportation engineering (english edition)}, vol.~3, no.~3, pp.~271--282,
  2016.

\bibitem{kampffmeyer2016semantic}
M.~Kampffmeyer, A.-B. Salberg, and R.~Jenssen, ``Semantic segmentation of small
  objects and modeling of uncertainty in urban remote sensing images using deep
  convolutional neural networks,'' in {\em Proceedings of the IEEE Conference
  on Computer Vision and Pattern Recognition Workshops}, pp.~1--9, 2016.

\bibitem{paisitkriangkrai2015effective}
S.~Paisitkriangkrai, J.~Sherrah, P.~Janney, V.-D. Hengel, {\em et~al.},
  ``Effective semantic pixel labelling with convolutional networks and
  conditional random fields,'' in {\em Proceedings of the IEEE Conference on
  Computer Vision and Pattern Recognition Workshops}, pp.~36--43, 2015.

\bibitem{Sherrah16}
J.~Sherrah, ``Fully convolutional networks for dense semantic labelling of
  high-resolution aerial imagery,'' {\em CoRR}, vol.~abs/1606.02585, 2016.

\bibitem{audebert2016semantic}
N.~Audebert, B.~Le~Saux, and S.~Lef{\`e}vre, ``Semantic segmentation of earth
  observation data using multimodal and multi-scale deep networks,'' in {\em
  Asian Conference on Computer Vision}, pp.~180--196, Springer, 2016.

\bibitem{MarmanisSWGDS16}
D.~Marmanis, K.~Schindler, J.~D. Wegner, S.~Galliani, M.~Datcu, and U.~Stilla,
  ``Classification with an edge: Improving semantic image segmentation with
  boundary detection,'' {\em CoRR}, vol.~abs/1612.01337, 2016.

\bibitem{wang2017gated}
H.~Wang, Y.~Wang, Q.~Zhang, S.~Xiang, and C.~Pan, ``Gated convolutional neural
  network for semantic segmentation in high-resolution images,'' {\em Remote
  Sensing}, vol.~9, no.~5, p.~446, 2017.

\bibitem{long2015fully}
J.~Long, E.~Shelhamer, and T.~Darrell, ``Fully convolutional networks for
  semantic segmentation,'' in {\em Proceedings of the IEEE Conference on
  Computer Vision and Pattern Recognition}, pp.~3431--3440, 2015.

\bibitem{yu2015multi}
F.~Yu and V.~Koltun, ``Multi-scale context aggregation by dilated
  convolutions,'' {\em arXiv preprint arXiv:1511.07122}, 2015.

\bibitem{he2015delving}
K.~He, X.~Zhang, S.~Ren, and J.~Sun, ``Delving deep into rectifiers: Surpassing
  human-level performance on imagenet classification,'' in {\em Proceedings of
  the IEEE international conference on computer vision}, pp.~1026--1034, 2015.

\bibitem{ioffe2015batch}
S.~Ioffe and C.~Szegedy, ``Batch normalization: Accelerating deep network
  training by reducing internal covariate shift,'' {\em arXiv preprint
  arXiv:1502.03167}, 2015.

\bibitem{ILSVRC15}
O.~Russakovsky, J.~Deng, H.~Su, J.~Krause, S.~Satheesh, S.~Ma, Z.~Huang,
  A.~Karpathy, A.~Khosla, M.~Bernstein, A.~C. Berg, and L.~Fei-Fei, ``{ImageNet
  Large Scale Visual Recognition Challenge},'' {\em International Journal of
  Computer Vision (IJCV)}, vol.~115, no.~3, pp.~211--252, 2015.

\bibitem{lovasz}
M.~Berman, A.~R. Triki, and M.~B. Blaschko, ``The lovasz-softmax loss: A
  tractable surrogate for the optimization of the intersection-over-union
  measure in neural networks,'' in {\em 2018 IEEE/CVF Conference on Computer
  Vision and Pattern Recognition}, pp.~4413--4421, IEEE, 2018.

\bibitem{KingmaB14adam}
D.~P. Kingma and J.~Ba, ``Adam: {A} method for stochastic optimization,'' {\em
  CoRR}, vol.~abs/1412.6980, 2014.

\bibitem{amsgrad2018}
S.~J. Reddi, S.~Kale, and S.~Kumar, ``On the convergence of {Adam} and
  beyond,'' {\em CoRR}, 2018.

\bibitem{ISPRS2018}
I.~S. for Photogrammetry and R.~S. (ISPRS), ``{2D Semantic Labeling Contest}.''
  online, 2018.

\bibitem{liuqh2018}
Q.~Liu, A.~Salberg, and R.~Jenssen, ``A comparison of deep learning
  architectures for semantic mapping of very high resolution images,'' in {\em
  IGARSS 2018 - 2018 IEEE International Geoscience and Remote Sensing
  Symposium}, pp.~6943--6946, July 2018.

\bibitem{salberg2017large}
A.-B. Salberg, {\O}.~D. Trier, and M.~Kampffmeyer, ``Large-scale mapping of
  small roads in lidar images using deep convolutional neural networks,'' in
  {\em Scandinavian Conference on Image Analysis}, pp.~193--204, Springer,
  2017.

\end{thebibliography}

\end{document}